\documentclass[10pt,twocolumn,letterpaper]{article}

\usepackage{cvpr}
\usepackage{times}
\usepackage{epsfig}
\usepackage{graphicx}
\usepackage{amsmath}
\usepackage{amssymb}

\usepackage{algpseudocode}
\usepackage{cases}
\usepackage{subfigure}
\usepackage{overpic}
\usepackage{multirow}
\usepackage{color}
\usepackage{array}
\usepackage{tabularx}
\usepackage[linesnumbered,ruled]{algorithm2e}
\usepackage{bbding}
\definecolor{mypink}{RGB}{219, 48, 122}

\usepackage{booktabs}
\usepackage{hhline}
\usepackage{makecell}
\usepackage{colortbl}
\usepackage[normalem]{ulem}
\usepackage{soul}
\usepackage{url}
\definecolor{Gray}{rgb}{0.7,0.7,0.7}

\newcolumntype{x}{>\small c}

\newcolumntype{o}{>\small L}

\usepackage[breaklinks=true,colorlinks,bookmarks=false]{hyperref}

\cvprfinalcopy 

\def\cvprPaperID{17} 

\ifcvprfinal\fi

\begin{document}

\title{Invariance Matters: Exemplar Memory for Domain Adaptive \\ Person Re-identification}

\author{Zhun Zhong$^{\textcolor{mypink}{1,2}}$, Liang Zheng$^{\textcolor{mypink}{3}}$, Zhiming Luo$^{\textcolor{mypink}{5}}$, Shaozi Li$^{\textcolor{mypink}{1}}$\thanks{Corresponding author (szlig@xmu.edu.cn). 
\newline \hspace*{0.16in} This work was done when Zhun Zhong (zhunzhong007@gmail.com) was a visiting student at University of Technology Sydney. Part of this work was done when Yi Yang (yee.i.yang@gmail.com) was visiting Baidu Research during his Professional Experience Program.
}~, Yi Yang$^{\textcolor{mypink}{2,4}}$ \\
 \small{\textcolor{mypink}{1} Cognitive Science Department, Xiamen University} \\
 \small{\textcolor{mypink}{2} Centre for Artificial
Intelligence, University of Technology Sydney}\\ 
\small{\textcolor{mypink}{3} Research School of Computer Science, Australian National University} \\
\small{\small{\textcolor{mypink}{4} Baidu Research} \space \space \textcolor{mypink}{5} Postdoc Center of Information and Communication Engineering, Xiamen University}}

\maketitle
\thispagestyle{empty}

\begin{abstract}

This paper considers the domain adaptive person re-identification (re-ID) problem: learning a re-ID model from a labeled source domain and an unlabeled target domain. Conventional methods are mainly to reduce feature distribution gap between the source and target domains.
However, these studies largely neglect the intra-domain variations in the target domain, which contain critical factors influencing the testing performance on the target domain. 
In this work, we comprehensively investigate into the intra-domain variations of the target domain and propose to generalize the re-ID model w.r.t three types of the underlying invariance, i.e., exemplar-invariance, camera-invariance and neighborhood-invariance. 
To achieve this goal, an exemplar memory is introduced to store features of the target domain and accommodate the three invariance properties. The memory allows us to enforce the invariance constraints over global training batch without significantly increasing computation cost.
Experiment demonstrates that the three invariance properties and the proposed memory are indispensable towards an effective domain adaptation system. 
Results on three re-ID domains show that our domain adaptation accuracy outperforms the state of the art by a large margin. Code is available at: \url{https://github.com/zhunzhong07/ECN}

\end{abstract}

\section{Introduction}

Person re-identification (re-ID) \cite{zheng2016personsurvery,zhong2017re,wu2019progressive,Li_2018_CVPR} is a cross-camera image retrieval task, which aims to find matched persons of a given query from the database. 
In spite of the impressive achievement of supervised learning in the re-ID community, learning a re-ID model that generalizes well on a target domain remains a challenge \cite{deng2018image,wang2018reid}.
Obtaining sufficient unlabeled data in the target domain is relatively easy, and yet it is difficult to learn a deep re-ID model without annotations.
This work considers the problem of unsupervised domain adaptation (UDA), where we are provided with labeled source domain and unlabeled target domain.
Our goal is to learn a discriminative representation for the target set.

\vspace{-.007in}
In the traditional setting of UDA, most methods are developed under the closed-set scenario, assuming that the source and target domains share entirely the same classes \cite{tzeng2017adversarial,ganin2015unsupervised}. 
However, this assumption cannot be applied to UDA in person re-ID, because the classes from the two domains are completely different. UDA in person re-ID is an open set problem \cite{busto2017open-set} which is more challenging than closed-set one. 
During UDA in person re-ID, it is improper to directly align the distributions of the source and target domains as in existing closed-set UDA methods. Instead, we should learn to well separate the unseen classes from the target domain.

\vspace{-.007in}
Recent advanced methods address the UDA problem in person re-ID mostly by reducing the gap between the source and target domains on the image-level \cite{deng2018image,wei2018person,Bak_2018_ECCV} or the attribute feature-level \cite{wang2018reid,lin2018multibmvc}.  
These methods only consider the overall inter-domain variations between the source and target domains, but ignore the intra-domain variations of the target domain. In fact, the target variations are critically influencing factors for person re-ID. 
In this study, we explicitly take into account the intra-domain variations of target domain and investigate three underlying invariances, \ie, exemplar-invariance, camera-invariance, and neighborhood-invariance, as described below.

\begin{figure*}[!t]
    \centering
    \includegraphics[width=0.97\linewidth]{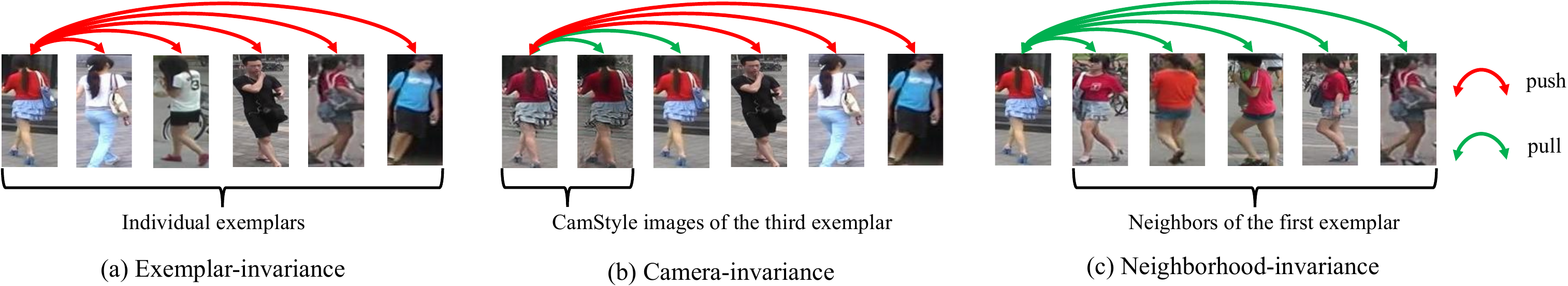}
    \caption{Examples of three underlying invariances. (a) Exemplar-invariance: an exemplar is enforced to be apart from others. (b) Camera-invariance: an exemplar and its camera-style transferred (CamStyle) images are encouraged to be close to each other, as well as CamStyle images should be far away from others. (c) Neighborhood-invariance: an exemplar and its neighbors are forced to be close to each other.}
    \label{fig:three_invariance}
\end{figure*}

\emph{First}, given a deep re-ID model trained on a labeled set, we observe that the top-ranked retrieval results always are more likely to be visually correlated to the query. 
A similar phenomenon is observed in image classification \cite{wu2018unsupervised}. 
This indicates that the deep re-ID model has learned the apparent similarity instead of semantic information from visual data. 
In reality, each person exemplar could differ significantly from other exemplars even belonged to the same identity. 
Thus, it is possible to enable the re-ID model to capture the apparent representation of person by learning to discriminate individual exemplars.
Based on this, we introduce the exemplar-invariance to learn apparent similarity on unlabeled target data by enforcing each person exemplar to be close to itself and far away from others. 
\emph{Second}, as a key influencing factor in person re-ID, camera-style variations \cite{zhong2018camera} might significantly change the appearance of person. Nevertheless, a person image generated by camera-style transfer still belongs to the original identity. Taking this into account, we enforce the camera-invariance \cite{Zhong_2018_ECCV} under the assumption that a person image and the corresponding camera-style transferred images should be close to each other.
\emph{Third}, suppose we are provided an appropriate re-ID model trained on the source and target domains. A target exemplar and its nearest-neighbors in the target set may probably have the same identity.  
Considering this trait, we present the neighborhood-invariance by encouraging an exemplar and its corresponding reliable neighbors to be close to each other. This helps us to learn a model that is more robust to overcome the image variations of the target domain, such as pose, view and background changes. Examples of these three invariances are shown in Fig.~\ref{fig:three_invariance}.

Based on the above aspects, we propose a novel unsupervised domain adaptation method for person re-ID. During the training process, an exemplar memory is introduced into the network to memorize the up-to-date representation of each exemplar of the target set. The memory enables us to enforce the invariance constraints over whole/global target training batch instead of the mini-batch. This helps us to effectively perform the invariance learning of the target domain during the network optimizing procedure. 

In summary, the contribution of this work is three-fold: 
\vspace{-.07in}
\begin{itemize}
\item We comprehensively study three underlying invariances of the target domain. Experiments show that these properties are indispensable for improving the transferable ability of re-ID models.
\vspace{-.09in}
\item We propose a memory module to effectively enforce the three invariance properties into the system. The memory helps us to take advantage of sample similarity over the global training set. With the memory, accuracy can be significantly improved, requiring very limited extra computation cost and GPU memory.
\vspace{-.03in}
\item Our method outperforms the state-of-the-art UDA approaches by a large margin on three large-scale datasets: Market-1501, DukeMTMC-reID and MSMT17.
\end{itemize}

\section{Related Work}

\textbf{Unsupervised domain adaptation.} 
An effective approach for addressing UDA is by aligning the feature distributions between the two domains. This alignment can be achieved by reducing the Maximum Mean Discrepancy (MMD) \cite{gretton2007kernel} between domains \cite{long2015learning,yan2017mind}, or training an adversarial domain-classifier \cite{bousmalis2016domain,tzeng2017adversarial} to encourage the features of the source and target domains to be indistinguishable. 
The above mentioned methods are designed under the assumption of the closed-set scenario, where the classes of the source and target domains are entirely identical. 
However, in practice, there are many scenarios that exist unknown classes in the target domain. The unknown-class samples from the target domain should not be aligned with the source domain. 
This task is introduced by Busto and Grall \cite{busto2017open-set}, referred as open set domain adaptation. To tackle this problem, Busto and Grall \cite{busto2017open-set} develop a method to learn a mapping from the source domain to the target domain by discarding unknown-class target samples. Recently, an adversarial learning framework \cite{saito2018open} is proposed to separate target samples into known and unknown classes, and reject unknown classes during feature alignment. In this paper, we study the problem of UDA in person re-ID, where the classes are totally different between the source and target domains. This is a more challenging open set problem.

\begin{figure*}[!t]
    \centering
    \includegraphics[width=0.98\linewidth]{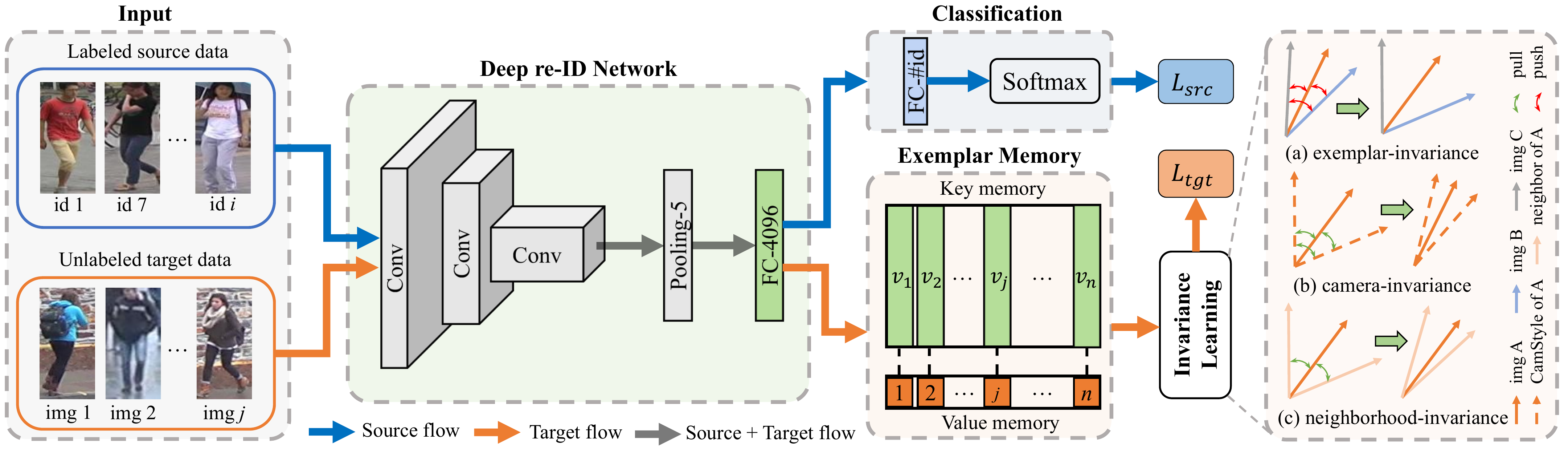}
    \caption{The framework of the proposed approach. During training, labeled source data and unlabeled target data are fed-forward into the deep re-ID network to obtain up-to-date representations. Subsequently, two components are designed to optimize the network with source data and target data, respectively. The first component is a classification module that calculates the cross-entropy loss for labeled source data. The second component is an exemplar memory module that saves the up-to-date features for target data and computes the invariance learning loss for unlabeled target data.}
    \label{fig:framework}
    \vspace{-.07in}
\end{figure*}

\textbf{Unsupervised person re-identification.}
The art supervised methods have made great achievement in person re-ID \cite{Li_2018_CVPR,sun2018beyond,zheng2019joint,sun2019dissecting}, relying on rich-labeled data and the success of deep networks \cite{Yawei2019Taking,resnet,dong2019search}. However, the performance may drop significantly when tested on an unseen dataset.
To address this problem, several methods use the labeled source domain to learn a deep re-ID model as an initialized feature extractor. Then, these methods learn a metric \cite{yu2017cross} or refine the re-ID model by unsupervised clustering \cite{fan2017pul} on the target domain. However, these methods do not take advantage of the labeled source data as a beneficial supervision during adapting procedure.
To overcome previous drawbacks, many domain adaptation approaches are developed to adapt the model with both labeled source domain and unlabeled target domain. These methods are mainly to reduce the domain shifts between datasets on image-level \cite{deng2018image,wei2018person,Bak_2018_ECCV} and attribute feature-level \cite{wang2018reid,lin2018multibmvc}. 
Despite their effectiveness, these methods largely ignore the intra-domain variations in target domain. 
Recently, Zhong \emph{et al.} \cite{Zhong_2018_ECCV} first propose a HHL method to learn camera-invariant network for the target domain. However, HHL overlooks the latent positive pairs in the target domain. This might lead the re-ID model to be sensitive to other variations in the target domain, such as pose and background variations.

\textbf{Difference from previous works.} Indeed, the three invariance properties and the memory module have been separately presented in existing works. However, our work is different from them. The exemplar-invariance and memory module have been presented in self-supervised learning \cite{wu2018unsupervised}, few-shot learning \cite{santoro2016meta,vinyals2016matching,wu2018improving} and supervised learning \cite{xiao2017joint}. Yet, we explore the feasibility of this idea in unsupervised domain adaptation and overcoming the variations in the target domain. 
The neighborhood-invariance is similar to deep association learning (DAL) \cite{chen2018deep}. A difference from DAL is that we design a soft classification loss to align the top-$k$ neighbors instead of calculating the triplet loss between the mutual top-1 neighbors.
Importantly, comparing with HHL \cite{Zhong_2018_ECCV} and DAL \cite{chen2018deep}, we comprehensively consider three invariance constraints. It is worthy of discovering the mutual benefit among the three invariance properties.

\section{The Proposed Method}

\textbf{Preliminary.} In the context of unsupervised domain adaptation (UDA) in person re-ID, we are provided with a fully labeled source domain \{$X_s, Y_s$\}, including $N_s$ person images. Each person image $x_{s,i}$ is associated with an identity $y_{s, i}$. The number of identities is $M$ for source domain. In addition, we are provided with an unlabeled target domain $X_t$, containing $N_t$ person images. The identity annotation of the target domain is not available. Our goal is to learn a transferable deep re-ID model using both labeled source domain and unlabeled target domain, which generalizes well on the target testing set.

\subsection{Overview of Framework}

The framework of our method is shown in Fig.~\ref{fig:framework}. In our model, the ResNet-50 \cite{resnet} pre-trained on ImageNet \cite{deng2009imagenet} is utilized as the backbone. Specifically, we keep the layers of ResNet-50 till the Pooling-5 layer as the base network and add a 4096-dimensional fully convolutional (FC) layer after Pooling-5 layer. The new FC layer is named FC-4096, followed by batch normalization \cite{ioffe2015batch}, ReLU \cite{nair2010relu}, Dropout \cite{srivastava2014dropout} and two components. The first component is a classification module for supervised learning with the labeled source data. It has an $M$-dimensional FC layer (named as FC-\#id) and a softmax activation function. We use the cross-entropy loss to calculate the loss for the source domain. The other component is an exemplar memory module for invariance learning with the unlabeled target data. The exemplar memory is served as a feature-storage that saves the up-to-date output of FC-4096 layer for each target image. We calculate the invariance learning loss of the target domain by estimating the similarities between the target samples within mini-batch and whole target samples saved in the exemplar memory.

\subsection{Supervised Learning for Source Domain}

Due to the identities of source images are available, we treat the training process of the source domain as a classification problem \cite{zheng2016personsurvery}. The cross-entropy loss is used to optimize the network, formulated as,
\begin{eqnarray}
    \begin{array}{l}
   \mathcal{L}_{src} = - \frac{1}{n_s} \sum\limits_{i=1}^{n_s} \log p(y_{s, i}|x_{s,i}),
   \label{cross-entropy-loss}
   \end{array}
\end{eqnarray}
where $n_s$ is the number of source images in a training batch. $p(y_{s, i}|x_{s,i})$ is the predicted probability that the source image $x_{s, i}$ belongs to identity $y_{s, i}$, which is obtained by the classification module.

The model trained using labeled source data produces a high accuracy on the same distributed testing set. However, the performance will deteriorate seriously when the testing set has a different distribution to the source domain. Next, we will introduce an exemplar memory based method to overcome this problem by considering the intra-domain variations of target domain in the training of network.

\subsection{Exemplar Memory}

In order to improve the generalization ability of the network on the target testing set, we propose to enforce invariance learning into the network by estimating the similarities between target images. 
To achieve this goal, we first construct an exemplar memory for storing the up-to-date features of all target images. The exemplar memory is a key-value structure \cite{xiao2017joint}, which has the key memory ($\mathcal{K}$) and the value memory ($\mathcal{V}$). 
In the exemplar memory, each slot stores the L2-normalized feature of FC-4096 in the key part, while storing the label in the value part. 
Given an unlabeled target data including $N_t$ images, we regard each image instance as an individual category. Thus, the exemplar memory contains $N_t$ slots, in which each slot storing the feature and label of a target image. 
In the initialization, we initialize the values of all the features in the key memory to zeros. 
For simplicity, we assign the corresponding indexes as the labels of target samples and store them in the value memory. For example, the class of \emph{i-th} target image in value memory is assigned to $\mathcal{V}$[$i$] = $i$. 
The labels in the value memory are fixed throughout training process. 
During each training iteration, for a target training sample $x_{t,i}$, we forward it through the deep reID network and obtain the L2-normalized feature of FC-4096, $f(x_{t,i})$. 
During the back-propagation, we update the feature in the key memory for the training sample $x_{t,i}$ through,
\begin{eqnarray}
  \begin{array}{l}
   \mathcal{K}[i] \leftarrow  \alpha \mathcal{K}[i] + (1 -  \alpha)  f(x_{t,i}),
   \label{memory-update}
   \end{array}
\end{eqnarray}
where $\mathcal{K}[i]$ is the key memory of image $x_{t,i}$ in $i$-\emph{th} slot. The hyper-parameter $\alpha$ $\in [0, 1]$ controls the updating rate. $\mathcal{K}[i]$ is then L2-normalized via $\mathcal{K}[i] \leftarrow \Vert \mathcal{K}[i] \Vert_2$.

\subsection{Invariance Learning for Target Domain}

The deep re-ID model trained with only source domain is usually sensitive to the intra-domain variations of the target domain. The variations are critical influencing factors for the performance. Therefore, it is necessary to consider the image variations of the target domain during transferring the knowledge from source domain to target domain. In this study, we investigate three underlying invariances of target data for UDA in person re-ID, \ie, exemplar-invariance, camera-invariance and neighborhood-invariance.

\textbf{Exemplar-invariance.} The appearance of each person image may be very different from others even shared the same identity. In other words, each person image can be close to itself while far away from others. Therefore, we enforce exemplar-invariance into the re-ID model by learning to distinguish individual person images. This allows the re-ID model to capture the apparent representation of person. To achieve this goal, we regard the $N_t$ target images as $N_t$ different classes, and classify each image into its own class. Given a target image $x_{t,i}$, we first compute the cosine similarities between the feature of $x_{t,i}$ and features saved in the key memory. Then, the predicted probability that $x_{t,i}$ belongs to class $i$ is calculated using softmax function,
\begin{eqnarray}
  \begin{array}{l}
  p(i|x_{t,i}) = \frac{\exp (\mathcal{K}[i]^\mathrm{T} f(x_{t,i}) / \beta)}{\sum_{j=1}^{N_t} \exp (\mathcal{K}[j]^\mathrm{T} f(x_{t,i}) / \beta)},
   \label{eq:probability}
   \end{array}
\end{eqnarray}
where $\beta \in (0, 1]$ is temperature fact that balances the scale of distribution.

The objective of exemplar-invariance is to minimize the negative log-likelihood over target training image, as
\begin{eqnarray}
    \begin{array}{l}
   \mathcal{L}_{ei} = - \log p(i|x_{t,i}).
   \label{exemplar-invariance}
   \end{array}
\end{eqnarray}

\textbf{Camera-invariance.} Camera style variation is an important factor in person re-ID. A person image may encounter with significant changes in appearance under different cameras. The re-ID model trained using labeled source data can capture the camera-invariance for source domain, but may suffer from the image variations caused by target cameras. Since the camera settings of the two domains will be very different.  
To overcome this problem, we propose to equip the network with camera-invariance \cite{Zhong_2018_ECCV} of target domain, based on the assumption that an image and its camera-style transferred counterparts should be close to each other. 
In this paper, we suppose the camera-ID of each image is known, since the camera-ID can be easily obtained when collecting person images from video sequences. Given the unlabeled target data, we consider each camera as a style domain and adopt StarGAN \cite{stargan} to train a camera style (CamStyle) transfer model \cite{zhong2018camera} for the target domain. With the learned CamStyle transfer model, each real target image collected from camera $c$ is augmented with $C-1$ images in the styles of other cameras while remaining the original identity. $C$ is the number of cameras in the target domain. 

To introduce the camera-invariance into the model, we regard that each real image and its style-transferred counterparts share the same identity. Thus, the loss function of camera-invariance is explained as,
\begin{eqnarray}
    \begin{array}{l}
   \mathcal{L}_{ci} = - \log p(i|{\hat{x}}_{t,i}),
   \label{camera-invariance}
   \end{array}
\end{eqnarray}
where ${\hat{x}}_{t,i}$ is a target sample randomly selected from the style-transferred images of $x_{t,i}$. In this way, images in different camera styles of the same sample are forced to be close to each other.

\textbf{Neighborhood-invariance.} For each target image, there may exist a number of positive samples in the target data. If we could exploit these positive samples in the training process, we are able to further improve the robustness of re-ID model in overcoming the variations of target domain. To achieve this objective, we first calculate the cosine similarities between $f(x_{t,i})$ and the features stored in the key memory $\mathcal{K}$. Then, we find the $k$-nearest neighbors of $x_{t,i}$ in $\mathcal{K}$ and define the indexes of them as $\mathcal{M}(x_{t,i}, k)$. $k$ is the size of $\mathcal{M}(x_{t,i}, k)$. The nearest one in $\mathcal{M}(x_{t,i}, k)$ is $i$.

We endow the neighborhood-invariance into the network under the assumption that the target image $x_{t,i}$ should belong to the classes of candidates in $\mathcal{M}(x_{t,i}, k)$. Thus, we assign the weight of the probability that $x_{t,i}$ belongs to the class $j$ as,
\begin{eqnarray}
  w_{i, j} =
  \begin{cases}
   \frac{1}{k}, &  j \neq i  \\ 
    1, & j = i 
  \end{cases}, \forall j \in \mathcal{M}(x_{t,i},k).
\label{weight}
\end{eqnarray}
The objective of neighborhood-invariance is formulated as a soft-label loss,
\begin{eqnarray}
    \begin{array}{c}
   \mathcal{L}_{ni} =  - \sum\limits_{j \neq i} w_{i, j} \log p(j|x_{t,i}), ~~ \forall j\in \mathcal{M}(x_{t,i},k).
   \label{neighborhood-invariance}
   \end{array}
\end{eqnarray}
Note that, to distinguish between exemplar-invariance and neighborhood-invariance, $x_{t,i}$ is not classified to its own class in Eq.~\ref{neighborhood-invariance}.

\textbf{Overall loss of invariance learning.} By jointly considering the exemplar-invariance, camera-invariance and neighborhood-invariance, the overall loss of invariance learning over target training images can be written as,
\begin{eqnarray}
    \begin{array}{l}
   \mathcal{L}_{tgt} =  - \frac{1}{n_t} \sum\limits_{i=1}^{n_t} \sum_{j} w_{i, j} \log p(j|x_{t,i}^{*}), 
   \label{target-loss}
   \end{array}
\end{eqnarray}
where $j \in \mathcal{M}(x_{t,i}^{*},k)$. $x_{t,i}^{*}$ is an image randomly sampled from the union set of $x_{t,i}$ and its camera style-transferred images. $n_t$ is the number of target images in the training batch. In Eq.~\ref{target-loss}, when $i=j$, we optimize the network with the exemplar-invariance learning and camera-invariance learning by classifying $x_{t,i}^{*}$ into its own class. When $i \neq j$, the network is optimized with the neighborhood-invariance learning by leading $x_{t,i}^{*}$ to be close to its neighbors in $\mathcal{M}(x_{t,i}^{*},k)$.

\subsection{Final Loss for Network}

By combining the losses of source and target domains, the final loss for the network is formulated as,
\begin{eqnarray}
    \begin{array}{l}
   \mathcal{L} =  (1 - \lambda) \mathcal{L}_{src} + \lambda \mathcal{L}_{tgt},
   \label{eq:final-loss}
   \end{array}
\end{eqnarray}
where $\lambda \in [0, 1]$ controls the importance of the source loss and the target loss. To this end, we introduce a loss function for UDA person re-ID, in which, the loss of source domain aims to maintain a basic representation for person. As well as, the loss of target domain attempts to take the knowledge from labeled source domain and incorporate the invariance properties of target domain into the network.

\subsection{Discussion on the Three Invariance Properties}
We analyze the advantage and disadvantage for each invariance. The exemplar-invariance enforces each exemplar away from each other. 
It is beneficial to enlarge the distance between exemplars from different identities. 
However, exemplars of the same identity will also be far apart, which is harmful to the system. 
On the contrast, neighborhood-invariance encourages each exemplar and its neighbors to be close to each other. 
It is beneficial to reduce the distance between exemplars of the same identity. 
However, neighborhood-invariance might also pull closer images of different identities, because we could not guarantee that each neighbor shares the same identity with the query exemplar. Therefore, there exists a trade off between exemplar-invariance and neighborhood-invariance, where the former aims to lead the exemplars from different identities to be far away while the latter attempts to encourage exemplars of the same identity to be close to each other. Camera-invariance has the similar effect as the exemplar-invariance and also leads the exemplar and its camera-style transferred samples to share the same representation.

\section{Experiment}

\subsection{Dataset} We evaluate the proposed method on three large-scale person re-identification (re-ID) benchmarks: Market-1501 \cite{zheng2015scalable}, DukeMTMC-reID \cite{ristani2016performance,zheng2017unlabeled} and MSMT17 \cite{wei2018person}. Performance is evaluated by the cumulative matching
characteristic (CMC) and mean Average Precision (mAP).


\begin{table}[!t]
\begin{center}
\newcolumntype{C}{>{\centering\arraybackslash}X}%
\newcolumntype{R}{>{\raggedleft\arraybackslash}X}%
\begin{tabularx}{\linewidth}{ c|CC|CC}
\hline
\multicolumn{1}{c|}{\multirow{2}{*}{$\beta$}} & \multicolumn{2}{c|}{Duke $\rightarrow$ Market-1501 } & \multicolumn{2}{c}{Market-1501 $\rightarrow$ Duke} \\ 
\cline{2-5}
& Rank-1&mAP&Rank-1&mAP\\
\hline
\hline
0.01 & 47.3 & 20.0&29.1&13.2\\
0.03 & 72.3 & 40.3& 59.7&35.7\\
0.05 & \textbf{75.1} & \textbf{43.0}&\textbf{63.3}&\textbf{40.4}\\
0.1 & 71.4 & 36.8&59.3&35.8\\
0.5 & 52.3 & 23.1&45.4&24.2\\
1.0 & 47.8 & 20.8&40.2&19.3\\
\hline
\end{tabularx}
\end{center}
\vspace{-.1in}
\caption{\label{tabel:temperature} Evaluation with different values of $\beta$ in Eq.~\ref{eq:probability}.}
\vspace{-.05in}
\end{table}

\subsection{Experiment Setting}

\textbf{Deep re-ID model.} We adopt ResNet-50 \cite{resnet} as the backbone of our model and initialize the model with the parameters pre-trained on ImageNet \cite{deng2009imagenet}. We fix the first two residual layers to save GPU memory. The input image is resized to 256 $\times 128$. During training, we perform random flipping, random cropping and random erasing \cite{zhong2017random} for data augmentation. The probability of dropout is set to 0.5. We train the model with a learning rate of 0.01 for ResNet-50 base layers and of 0.1 for the others in the first 40 epochs. The learning rate is divided by 10 for the next 20 epochs. The SGD optimizer is used to train the model. We set the mini-batch size to 128 for both source images and target images. We initialize the updating rate of key memory $\alpha$ to 0.01 and increase $\alpha$ linearly with the number of epochs, \ie, $\alpha=0.01 \times epoch$. Without specification, we set the temperature fact $\beta = 0.05$, number of candidate positive samples $k=6$ and weight of losses $\lambda=0.3$. We train the model with exemplar-invariance and camera-invariance learning at the first 5 epochs and add the neighborhood-invariance learning for the rest epochs. In testing, we extract the L2-normalized output of Pooling-5 layer as the image feature and adopt the Euclidean distance to measure the similarities between query and gallery images.

\textbf{Baseline.} We set the model as the \emph{baseline} when trained the network using only the classification component.

\subsection{Parameter Analysis}

We first analyze the sensitivities of our approach to three important hyper-parameters, \ie, the temperature fact $\beta$, the weight of losses $\lambda$, and the number of candidate positive samples $k$. By default, we vary the value of one parameter and keep the others fixed.

\textbf{Temperature fact $\beta$}. In Table~\ref{tabel:temperature}, we investigate the effect of the temperature fact $\beta$ in Eq.~\ref{eq:probability}. Using a lower value for $\beta$ leads to a lower entropy, which commonly achieves better results. However, the network does not converge if the temperature fact is too low, \emph{e.g.}, $\beta = 0.01$. The best results are produced when $\beta$ is around 0.05.

\begin{figure}[!t]
\centering
\includegraphics[width=\linewidth]{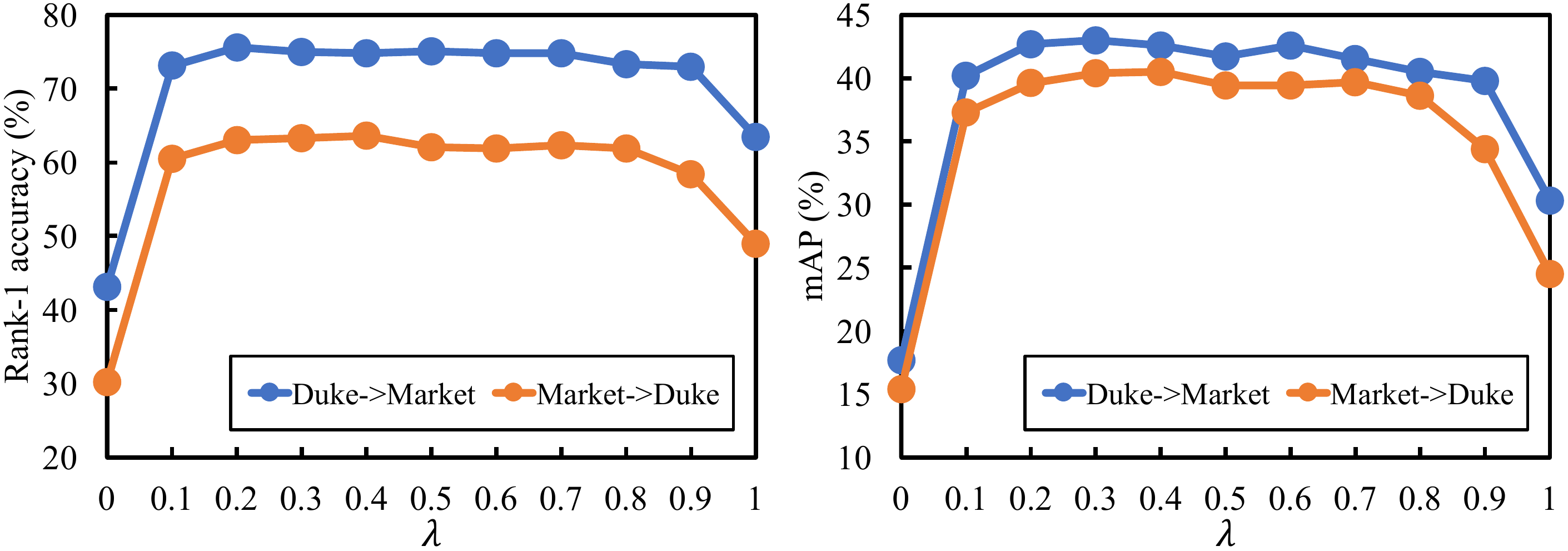}
\caption{Evaluation with different values of $\lambda$ in Eq.~\ref{eq:final-loss}.}
\label{fig:labmda}
\end{figure}

\textbf{The weight of source and target losses $\lambda$.} In Fig.~\ref{fig:labmda} we compare different values of $\lambda$ in Eq.~\ref{eq:final-loss}. When $\lambda = 0$, our method reduces to the baseline that trained the model only with labeled source data. It is clearly shown that, when considering invariance learning for target domain ($\lambda>0$), our approach significantly improves the baseline at all values. It is worth noting that our approach outperforms the baseline by a large margin even trained the model using only unlabeled target data ($\lambda=1$). This demonstrates the effectiveness of our approach and the importance of overcoming the variations in target domain. When $\lambda$ is between 0.3 to 0.8, our result is impacted just marginally and the best results are obtained. This shows that our method is insensitive to $\lambda$ in an appropriate range.

\begin{figure}[!t]
\centering
\includegraphics[width=\linewidth]{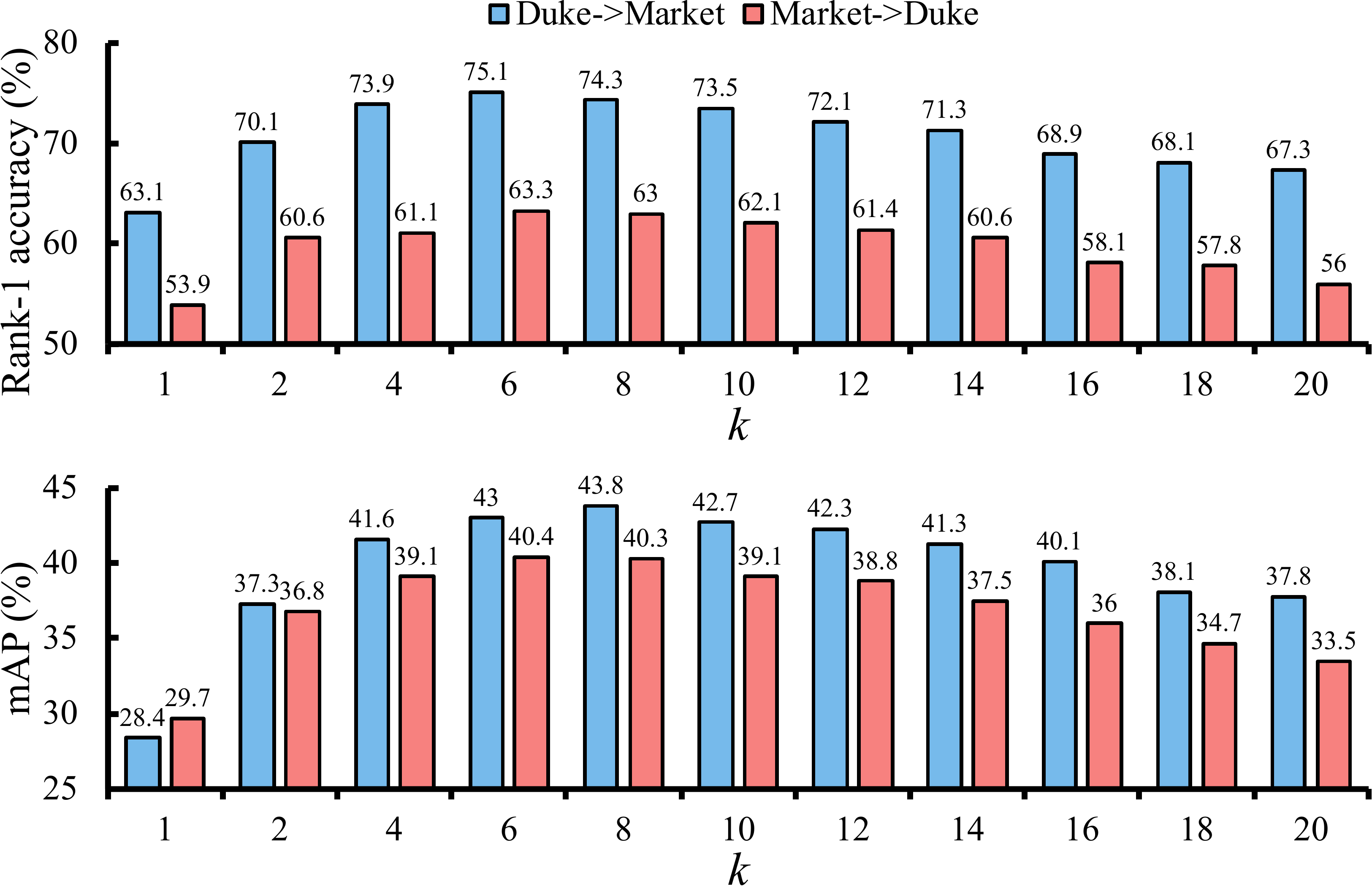}
\caption{Evaluation with different number of candidate positive samples in neighborhood-invariance learning.}
\label{fig:different_k}
\end{figure}

\begin{table*}[!t]
\begin{center}
\newcolumntype{C}{>{\centering\arraybackslash}X}%
\newcolumntype{R}{>{\raggedleft\arraybackslash}X}%
\begin{tabularx}{\linewidth}{ l||CCCCCC||CCCCCC}
\hline
\multicolumn{1}{l||}{\multirow{2}{*}{Methods}}&\multicolumn{6}{c||}{Market-1501}&\multicolumn{6}{c}{DukeMTMC-reID}\\
\cline{2-13}
\multicolumn{1}{c||}{}&Src.&R-1&R-5&R-10&R-20&mAP&Src.&R-1&R-5&R-10&R-20&mAP\\
\hline
\hline
Supervised Learning&N/A& 87.6& 95.5& 97.2& 98.3&69.4& N/A& 75.6& 87.3& 90.6& 92.9&57.8\\
\hline
Source Only& \multirow{5}{*}{\rotatebox{90}{DukeMTMC}} & 43.1& 58.8& 67.3& 74.3& 17.7&\multirow{5}{*}{\rotatebox{90}{Market-1501}}  &28.9&44.0&50.9&57.5&14.8\\
Ours w/ E& &48.7&67.4&74.0&80.2&21.0& &34.2&51.3&58&64.2&18.7\\
Ours w/ E+C& &63.1&79.1&84.6&89.1&28.4& &53.9&70.8&76.1&80.7&29.7\\
Ours w/ E+N& &58.0&69.9&75.6&80.4&27.7&&39.7&53.0&58.1&62.9&23.6\\
Ours w/ E+C+N& &\textbf{75.1}&\textbf{87.6}&\textbf{91.6}&\textbf{94.5} & \textbf{43.0}&&\textbf{63.3}&\textbf{75.8}&\textbf{80.4}& \textbf{84.2}&\textbf{40.4}\\
\hline
\end{tabularx}
\end{center}
\vspace{-.1in}
\caption{\label{tabel:ablation} Methods comparison when tested on Market-1501 and DukeMTMC-reID. \textbf{Supervised Learning}: Baseline model trained with labeled target data. \textbf{Source Only}: Baseline model trained with only labeled source data. \textbf{E}: Exemplar-invariance. \textbf{C}: Camera-invariance. \textbf{N}: Neighborhood-invariance. \textbf{Src.}: Source domain.}
\end{table*}

\textbf{Number of positive samples $k$.} In Fig.~\ref{fig:different_k}, we show the results of using different number of positive samples in neighborhood-invariance learning. When $k=1$, our approach reduces to the model trained with exemplar-invariance and camera-invariance learning. When adding neighborhood-invariance learning into the system ($k>1$), our results achieve consistent improvement. The rank-1 accuracy and mAP first improve with the increase of $k$ and achieve best results when $k$ is between 6 to 8. Assigning a too large value to $k$ reduces the results. This is because an excess of false positive samples may include during neighborhood-invariance learning, which could have deleterious effects on performance.

According to the analysis above, we set $\beta = 0.05$, $\lambda=0.3$ and $k=6$ in the following experiment.

\subsection{Evaluation}

\textbf{Performance of baseline.} Table~\ref{tabel:ablation} reports the results of the baseline. When trained with labeled target training set and tested on the target testing set, the baseline (called \emph{Supervised Learning}) achieves high accuracy. However, we observe a serious drop in performance when the baseline is trained using labeled source set only (called \emph{Source Only}) and directly applied to the target testing set. For example, when tested on Market-1501, the baseline trained on labeled Market-1501 training set achieves a rank-1 accuracy of 87.6\%. However, the rank-1 accuracy declines to 43.1\% when trained the baseline on labeled DukeMTMC-reID training set. A similar drop can be observed when tested on DukeMTMC-reID. This decline in accuracy is mainly caused by the domain shifts between datasets.

\textbf{Ablation experiment on invariance learning.} To investigate the effectiveness of the proposed invariance learning for target domain, we conduct ablation studies in Table~\ref{tabel:ablation}. First, we show the effect of exemplar-invariance learning by adding exemplar-invariance learning into the baseline. As shown in Table~\ref{tabel:ablation}, ``Ours w/ E'' consistently improves the results over baseline (Source Only). Specifically, the rank-1 accuracy improves from 43.1\% to 48.7\% and 28.9\% to 34.2\% when tested on Market-1501 and DukeMTMC-reID, respectively. This demonstrates that exemplar-invariance learning is an effective way to improve the discrimination of person descriptors for the target domain.

Next, we validate the effectiveness of camera-invariance learning over the model trained with exemplar-invariance learning (Ours w/ E). In Table~\ref{tabel:ablation}, we observe significant improvement when adding camera-invariance learning into the system. For example, ``Ours w/ E+C'' achieves a rank-1 accuracy of 63.1\% when regarding DukeMTMC-reID as source domain and tested on Market-1501. This is higher than ``Ours w/ E'' by 14.4\% in rank-1 accuracy. The improvement demonstrates that the image variations caused by target cameras severely impact the performance in testing set. Injecting camera-invariance learning into the model could effectively improve the robustness of the system to camera style variations.

We also evaluate the effect of neighborhood-invariance learning. As reported in Table~\ref{tabel:ablation}, ``Ours w/ E+N'' consistently improves the results of ``Ours w/ E''. Using exemplar-invariance and neighborhood-invariance during training, the model (Ours w/ E+N) has 39.7\% rank-1 accuracy and 23.6\% mAP when using Market-1501 as source domain and tested on DukeMTMC-reID. This increases the results of ``Ours w/ E'' by 5.5\% in rank-1 accuracy and by 4.9\% in mAP, respectively. Furthermore, when integrating neighborhood-invariance learning into a better model (Ours w/ E+C), our approach would gain more improvement in performance. For example, ``Ours w/ E+C+N'' achieves rank-1 accuracy of 75.1\% when regarding DukeMTMC-reID as source domain and tested on Market-1501, improving the rank-1 accuracy of ``Ours w/ E+C'' by 12\%. Similar improvement is observed when tested on DukeMTMC-reID. This is because that more reliable positive samples would be mined from unlabeled target set by integrating neighborhood-invariance learning into a more discriminative model.

\begin{table}[!t]
\small
\begin{center}
\newcolumntype{C}{>{\centering\arraybackslash}X}%
\newcolumntype{R}{>{\raggedleft\arraybackslash}X}%
\begin{tabularx}{\linewidth}{ l|C|c||c}
\hline
\multirow{2}{*}{Method} & \multicolumn{3}{c}{DukeMTMC-reID $\rightarrow$ Market-1501} \\
\cline{2-4}
& R-1 & Time (mins) & Memory (MB) \\
\hline
Ours w/ mini-batch & 67.2 & $\approx$ 59.3 & $\approx$5,000 \\
Ours w/ memory & \textbf{75.1} & $\approx$ 60.6 & $\approx$5,260 \\
\hline
\end{tabularx}
\end{center}
\vspace{-.12in}
\caption{\label{tabel:1} Computational cost analysis of the exemplar memory.}
\end{table}

\noindent \textbf{The benefit of the exemplar memory.} We use the proposed exemplar memory and the mini-batch to implement the proposed invariance learning, respectively. For mini-batch based method, input samples are composed of the target samples, corresponding CamStyle samples and corresponding $k$-nearest neighbors. As shown in Table \ref{tabel:1}, the exemplar memory based method clearly outperforms the mini-batch based method. It is noteworthy that using the exemplar memory introduces limited additional training time ($\approx$ + 1.3 mins) and GPU memory ($\approx$ + 260 MB) compared to using the mini-batch.

\begin{table*}[!t]
\begin{center}
\newcolumntype{C}{>{\centering\arraybackslash}X}%
\newcolumntype{R}{>{\raggedleft\arraybackslash}X}%
\begin{tabularx}{0.9\linewidth}{ l|CCCC|CCCC }
\hline
\multicolumn{1}{l|}{\multirow{2}{*}{Methods}} &\multicolumn{4}{c}{Market-1501}&\multicolumn{4}{c}{DukeMTMC-reID}\\
\cline{2-9}
\multicolumn{1}{c|}{}&R-1&R-5&R-10&mAP&R-1&R-5&R-10&mAP\\
\hline
\hline
LOMO \cite{liao2015lomo}&27.2&41.6&49.1&8.0&12.3&21.3&26.6&4.8\\
Bow \cite{zheng2015scalable}&35.8&52.4&60.3&14.8&17.1&28.8&34.9&8.3\\
\hline
UMDL \cite{peng2016unsupervised}&34.5&52.6&59.6&12.4&18.5&31.4&37.6&7.3\\
PTGAN \cite{wei2018person}& 38.6 & - & 66.1 & - & 27.4 & - & 50.7 & -\\
PUL \cite{fan2017pul}&45.5&60.7&66.7&20.5&30.0&43.4&48.5&16.4\\
SPGAN \cite{deng2018image}& 51.5&70.1&76.8& 22.8& {41.1}& {56.6}& {63.0}& {22.3}\\
CAMEL \cite{yu2017cross}&54.5&-&-&26.3&-&-&-&-\\
MMFA \cite{lin2018multibmvc}& 56.7&75.0&81.8& 27.4&45.3& 59.8& 66.3& 24.7\\
{SPGAN+LMP} \cite{deng2018image}& 57.7&75.8&82.4&26.7& 46.4&62.3&68.0&26.2\\
TJ-AIDL \cite{wang2018reid}& 58.2&74.8&81.1& 26.5& 44.3& 59.6& 65.0& 23.0\\
CamStyle \cite{zhong2019camstyle}&58.8& 78.2& 84.3&27.4&48.4&62.5& 68.9&25.1\\
HHL \cite{Zhong_2018_ECCV}&62.2&78.8&84.0&31.4 & 46.9&61.0&66.7&27.2\\
\hline
Ours (ECN)&\textbf{75.1}&\textbf{87.6}&\textbf{91.6}&\textbf{43.0}&\textbf{63.3}&\textbf{75.8}&\textbf{80.4}&\textbf{40.4}\\
\hline
\end{tabularx}
\end{center}
\vspace{-.1in}
\caption{\label{tabel:sota-m-d} Unsupervised person re-ID performance comparison with state-of-the-art methods on Market-1501 and DukeMTMC-reID.}
\vspace{.08in}
\end{table*}

\subsection{Comparison with State-of-the-art Methods}

We compare our approach with state-of-the-art unsupervised learning methods when tested on Market-1501, DukeMTMC-reID and MSMT17.

Table~\ref{tabel:sota-m-d} reports the comparisons when tested on Market-1501 and DukeMTMC-reID. We use DukeMTMC-reID as the source set when tested on Market-1501 and vice versa. We compare with two hand-crafted feature based methods without transfer learning: LOMO \cite{liao2015lomo} and BOW \cite{zheng2015scalable}, three unsupervised methods that use a labeled source data to initialize the model but ignore the labeled source data during learning feature for target domain: CAMEL \cite{yu2017cross}, UMDL \cite{peng2016unsupervised} and PUL \cite{fan2017pul}, and six unsupervised domain adaptation approaches: PTGAN \cite{wei2018person}, SPGAN \cite{deng2018image}, MMFA \cite{lin2018multibmvc}, TJ-AIDL \cite{wang2018reid}, CamStyle \cite{zhong2019camstyle}, HHL \cite{Zhong_2018_ECCV}. We first compare with hand-crafted feature based methods which do not require learning on neither labeled source set nor unlabeled target set. These two hand-crated features have demonstrated the effectiveness on small datasets, but fail to produce competitive results on large-scale datasets. For example, the rank-1 accuracy of LOMO is 12.3\% when tested on DukeMTMC-reID. This is much lower than transferring learning based methods. Next, we compare with three unsupervised methods. Benefit from initializing model from the labeled source data and learning with unlabeled target data, the results of these three unsupervised approaches are commonly superior to hand-crafted methods. For example, PUL obtains rank-1 accuracy of 45.5\% when using DukeMTMC-reID as source set and tested on Market-1501, surpassing BOW by 9.7\% in rank-1 accuracy. Compare to state-of-the-art domain adaptation approaches, our approach clearly outperforms them by a large margin on both datasets. Specifically, our method achieves \textbf{rank-1 accuracy = 75.1\%} and \textbf{mAP = 43.0\%} when using DukeMTMC-reID as source set and tested on Market-1501, and, obtains \textbf{rank-1 accuracy = 63.3\%} and \textbf{mAP = 40.4\%} when using Market-1501 as source set and tested on DukeMTMC-reID.  The rank-1 accuracy is 12.9\% higher  and 16.4\% higher than current best results (HHL \cite{Zhong_2018_ECCV}) when tested on Market-1501 and DukeMTMC-reID, respectively.

\begin{table}[!t]
\begin{center}
\newcolumntype{C}{>{\centering\arraybackslash}X}%
\newcolumntype{R}{>{\raggedleft\arraybackslash}X}%
\begin{tabularx}{\linewidth}{l|c|CCCC}
\hline
\multicolumn{1}{l|}{\multirow{2}{*}{Methods}}& \multirow{2}{*}{Src.} &\multicolumn{4}{c}{MSMT17}\\
\cline{3-6}
& &\small{R-1}&\small{R-5}&\small{R-10}&\small{mAP}\\
\hline
\hline
PTGAN \cite{wei2018person}& Market& 10.2&-& 24.4& 2.9\\
Ours (ECN)& Market&\textbf{25.3}&\textbf{36.3}&\textbf{42.1}&\textbf{8.5}\\
\hline
PTGAN \cite{wei2018person}& Duke&11.8&-&27.4& 3.3\\
Ours (ECN)& Duke& \textbf{30.2}&\textbf{41.5}&\textbf{46.8}&\textbf{10.2}\\
\hline
\end{tabularx}
\end{center}
\vspace{-.1in}
\caption{\label{tabel:MSMT17} Performance evaluation when tested on MSMT17.}
\end{table}

We also evaluate our approach on a larger and more challenging dataset, \ie, MSMT17. Since it is a newly released dataset, there is only one unsupervised method (PTGAN \cite{wei2018person}) reported on MSMT17. As shown in Table~\ref{tabel:MSMT17}, our approach clearly surpasses PTGAN when using Market-1501 and DukeMTMC-reID as source domains. For example, our method produces \textbf{rank-1 accuracy = 30.2\%} and \textbf{mAP = 10.2\%} when using DukeMTMC-reID as source set. This is higher than PTGAN by 18.4\% in rank-1 accuracy and by 6.9\% in mAP.

\section{Conclusion}

In this paper, we propose an exemplar memory based unsupervised domain adaptation (UDA) method for person re-ID task. With the exemplar memory, we can directly evaluate the relationships between target samples. And thus we could effectively enforce the underlying invariance constraints of the target domain into the network training process. Experiment demonstrates the effectiveness of the invariance learning for improving the transferable ability of deep re-ID model. Our approach produces a new state of the art in UDA accuracy on three large-scale domains. \\

{\noindent\textbf{Acknowledgements} This work is supported by the National Nature Science Foundation of China (No. 61876159, No. 61806172, No. 61572409, No. U1705286 \& 61571188), Fujian Province 2011Collaborative Innovation Center of TCM Health Management, Collaborative Innovation Center of Chinese Oolong Tea Industry-Collaborative Innovation Center (2011) of Fujian Province. Zhun Zhong thanks Wenjing Li for encouragement.}

{\small
\bibliographystyle{ieee}
\bibliography{egbib}
}

\end{document}